\documentclass[conference]{IEEEtran}
\IEEEoverridecommandlockouts
\usepackage{cite}

\usepackage{amsmath,amssymb,amsfonts,mathtools,bm}
\usepackage{amsthm}
\usepackage{algorithm}
\usepackage{algorithmic}
\usepackage{graphicx}
\usepackage{textcomp}
\usepackage{xcolor,float}
\usepackage{tabularx}
\usepackage{textcomp}
\usepackage{diagbox}
\usepackage{svg}
\usepackage{booktabs}
\usepackage{subcaption}
\usepackage{multirow}

\allowdisplaybreaks
\renewcommand{\baselinestretch}{1}

\begin{document}
    
% \title{{ALWNN Empowered Automatic Modulation Classification: Conquering Complexity and Scarce Sample Conditions}}
\title{\fontsize{20pt}{12pt}\selectfont ALWNN Empowered Automatic Modulation Classification: Conquering Complexity and Scarce Sample Conditions}

\author{
\IEEEauthorblockN{
Yunhao Quan\IEEEauthorrefmark{1},
Chuang Gao\IEEEauthorrefmark{2},
Nan Cheng\IEEEauthorrefmark{1},
Zhijie Zhang\IEEEauthorrefmark{1},
Zhisheng Yin\IEEEauthorrefmark{1},
Wenchao Xu\IEEEauthorrefmark{1}, and 
Danyang Wang\IEEEauthorrefmark{1}\\
}
\IEEEauthorblockA{
\IEEEauthorrefmark{1}School of Telecommunications Engineering,
Xidian University, Xi'an, China\\
\IEEEauthorrefmark{2}CSSC International Engineering Co., Ltd., Beijing 100121, China; \\
Email: \{qyh, xcwang\_1\}@stu.xidian.edu.cn; \{zsyin, dywang\}@xidian.edu.cn}; dr.nan.cheng@ieee.org;  gaochuang@csic602.com.cn}

\maketitle

\IEEEdisplaynontitleabstractindextext

\IEEEpeerreviewmaketitle

\begin{abstract}
In Automatic Modulation Classification (AMC), deep learning methods have shown remarkable performance, offering significant advantages over traditional approaches and demonstrating their vast potential. Nevertheless, notable drawbacks, particularly in their high demands for storage, computational resources, and large-scale labeled data, which limit their practical application in real-world scenarios. To tackle this issue, this paper innovatively proposes an automatic modulation classification model based on the Adaptive Lightweight Wavelet Neural Network (ALWNN) and the few-shot framework (MALWNN). The ALWNN model, by integrating the adaptive wavelet neural network and depth separable convolution, reduces the number of model parameters and computational complexity. The MALWNN framework, using ALWNN as an encoder and incorporating prototype network technology, decreases the model's dependence on the quantity of samples.
Simulation results indicate that this model performs remarkably well on mainstream datasets. Moreover, in terms of Floating Point Operations Per Second (FLOPS) and Normalized Multiply - Accumulate Complexity (NMACC), ALWNN significantly reduces computational complexity compared to existing methods. This is further validated by real-world system tests on USRP and Raspberry Pi platforms. Experiments with MALWNN show its superior performance in few-shot learning scenarios compared to other algorithms.

\end{abstract}


\section{Introduction}
% The rapid advancement of wireless technologies such as the Internet of Things (IoT) and the fifth generation of telecommunications technologies (5G) has led to an increasingly complex electromagnetic environment, exacerbating the scarcity of spectrum resources\cite{Background}. In this context, wireless communication can be divided into two modes: cooperative and non-cooperative communication. In cooperative communication, involved parties negotiate parameters and protocols\cite{VTC}, effectively addressing spectrum scarcity challenges through strategies such as spectrum sharing, cooperation, dynamic spectrum allocation, and multiple access techniques\cite{TCCN}. While cooperative communication is predominant in civilian communication, in areas such as wireless spectrum sharing and Unmanned Aerial Vehicle (UAV) communication, it is impossible for communication parties to negotiate modulation schemes, leading to situations where communication parties do not know the modulation schemes in advance. This can result in low communication efficiency and low spectrum utilization. Automatic Modulation Classification (AMC) is capable of autonomously discerning the modulation type in incoming signals, without pre-existing knowledge of the transmitted signal, thereby providing an effective solution to the previously stated problems. Through AMC, spectrum awareness capability can be enhanced, thereby alleviating the challenge of spectrum scarcity in non-cooperative communication scenarios and improving communication efficiency. 
The rapid evolution of modern wireless communication technologies has led to increasingly sophisticated modulation schemes in wireless channels\cite{Background}, while burgeoning user demands and exponential growth in data transmission have created congested electromagnetic environments that intensify spectrum scarcity\cite{Radiodiff}. This paradigm shift necessitates robust automatic modulation classification (AMC) solutions capable of rapidly identifying signal modulation types without prior transmitter knowledge. As an indispensable bridge between signal detection and demodulation, AMC plays a vital role in both civilian and military applications including cognitive radio networks, electronic warfare systems, adaptive modulation architectures, and spectrum monitoring infrastructures. The critical need for efficient AMC implementations stems from their fundamental position in enabling intelligent signal processing across next-generation communication ecosystems.

% Typically, AMC methods can be bifurcated into two primary categories: methods based on likelihood\cite{likehood} and those based on features\cite{feature}, both of which rely on acquiring channel state information (CSI). Likelihood-based methods directly depend on the CSI and compute the likelihood function under the assumption of known CSI to find the optimal Bayesian estimate. However, due to the inability to obtain CSI through pilot signals in non-cooperative communication scenarios, their adaptability is limited. Feature-based methods rely on the distribution of CSI extracted from the data and do not directly depend on CSI. Feature-based methods consist of feature extraction and classification. During the feature extraction phase, expert systems meticulously derive a variety of artificial features from the data, encompassing instantaneous as well as statistical features. Instantaneous features encompass elements like instantaneous phase, frequency, and amplitude. On the other hand, statistical features involve higher-order cumulants and moments. Following this, the derived features are utilized to construct robust machine learning models, such as decision trees, support vector machines, and artificial neural networks. Since the distribution of signal data's CSI also changes in non-cooperative communication scenarios, and the feature extraction part relies on expert knowledge, feature-based methods also have significant limitations in adaptability.
Conventional AMC approaches are systematically categorized into two principal paradigms: likelihood-based (LB)\cite{likehood} and feature-based (FB) \cite{feature}methodologies. The LB framework formulates AMC as a hypothesis testing problem through Bayesian estimation derived from signal likelihood functions, yet demonstrates critical constraints including susceptibility to channel impairments such as frequency offsets and multipath fading, prohibitive computational complexity causing impractical latency exceeding one second per classification, and fundamental incompatibility with real-time processing requirements. In contrast, FB techniques employ a dual-stage architecture comprising expert-designed feature extraction followed by machine learning classification. The feature extraction stage focuses on deriving instantaneous characteristics such as phase variations combined with statistical descriptors including higher-order cumulants, while the classification phase utilizes models ranging from decision trees to neural networks. However, conventional LB/FB methods exhibit operational limitations through rigid feature engineering that impedes adaptation to dynamic channel conditions, substantial dependence on domain-specific prior knowledge, and suboptimal efficiency in evolving electromagnetic environments. This technical landscape necessitates next-generation AMC solutions capable of balancing computational efficiency with environmental adaptability while maintaining classification fidelity.

% Deep learning (DL), with its powerful feature extraction capabilities and data-driven nature, is considered key to addressing the challenges in AMC. It does not rely on CSI but rather on the modulation scheme of the signals. O’shea et al. \cite{OSHEA} proposed a convolutional neural network (CNN) model for AMC, trained on time-domain data and employing a softmax classifier for feature classification, outperforming traditional methods reliant solely on handcrafted features. F. Meng et al.\cite{FMENG} proposed a CNN for AMC, based on a two-stage training strategy and transfer learning. However, these CNN-based approaches primarily focus on local spatial features, neglecting temporal correlations. To fully exploit both temporal and spatial features, West et al.\cite{CLDNN} combined recurrent neural networks (RNN) with CNN, proposing a Convolutional Long Short-Term Deep Neural Network (CLDNN) that significantly improved classification accuracy by capturing spatiotemporal features. Xu et al.\cite{MCLDNN} addressed the issue of insufficient spatial feature extraction in the convolutional part of CLDNN by introducing a Multi-Channel Convolutional Long Short-Term Deep Neural Network (MCLDNN). 

Deep learning (DL), with its hierarchical feature extraction capabilities and data-driven learning paradigm, has become fundamental in addressing AMC challenges. Unlike conventional approaches requiring prior knowledge of transmission parameters, DL-based methods autonomously learn discriminative patterns directly from raw signal representations. O'Shea \textit{et al.} \cite{OSHEA} pioneered the application of convolutional neural networks (CNNs) to AMC, demonstrating superior performance over handcrafted-feature methods through end-to-end learning from time-domain in-phase/quadrature (IQ) samples. Subsequently, Meng \textit{et al.} \cite{FMENG} enhanced CNN robustness via a dual-phase training strategy incorporating transfer learning. Nevertheless, these CNN-centric approaches primarily exploit localized spatial features while disregarding temporal dependencies in signal sequences. To comprehensively model spatiotemporal characteristics, West \textit{et al.} \cite{CLDNN} developed the convolutional long short-term deep neural network (CLDNN), synergistically integrating recurrent neural networks (RNNs) with CNNs. Further advancing this architecture, Xu \textit{et al.} \cite{MCLDNN} proposed the multi-channel CLDNN (MC-LDNN), which augments spatial feature extraction through parallelized convolutional pathways.

While DL-based AMC demonstrates promising capabilities, two domain-specific barriers hinder practical deployment\cite{EDGE}. First, dynamic spectrum sharing disrupts signal statistical regularity, adversarial interference obscures modulation fingerprints, and time-varying channels invalidate static training assumptions \cite{challenge}. These interacting phenomena create compounded data reliability challenges under embedded platform constraints. Second, inherent architectural complexity of DL-based AMC methods induces computational overload incompatible with resource-constrained edge devices\cite{CODE}, causing prohibitive latency and power consumption for real-time systems \cite{challenge1}. This dual exigency demands lightweight learning frameworks that synergistically address environmental uncertainty and hardware efficiency through computational paradigm innovation while preserving classification robustness\cite{lightweight}.

This paper proposes ALWNN, a lightweight neural network for AMC that integrates adaptive wavelet transformations and depthwise separable convolutions. The architecture achieves multi-level feature extraction while reducing computational complexity and parameter count by one to two orders of magnitude. For few-shot scenarios, we introduce MALWNN, a prototype network-based framework leveraging ALWNN as its encoder, which exhibits superior data efficiency compared to existing methods. Practical deployments on USRP platforms and Raspberry Pi 4B systems validate the framework’s effectiveness. The main contributions of this paper are summarized as follows.

\begin{enumerate}
    
    \item A lightweight model architecture is proposed for efficient and accurate signal modulation classification, incorporating depthwise separable convolutions to reduce model complexity and enhance computational efficiency.
    \item To overcome the dependence of traditional DL methods on a large amount of labeled data, MALWNN is proposed to enhance the performance of the model under the condition of small-sample labeled data.

    \item The AWLNN architecture achieved superior performance on benchmark datasets, maintaining comparable accuracy to conventional methods while reducing parameters and computational complexity by 1-2 orders of magnitude. Hardware implementations on USRP and Raspberry Pi 4B platforms further verified these efficiency improvements.
    \item A large number of evaluations based on the RadioML2018.01a dataset for the proposed MALWNN framework have shown that MALWNN achieves better accuracy than existing methods under a small number of samples.

\end{enumerate}

The remainder of this paper is organized as follows. The related work is introduced in Section II. In Section III, we introduce the system model and conduct problem simulations. In Section IV, the proposed ALWNN model and MALWNN architecture will be introduced. In Section V, the performance of the proposed ALWNN algorithm and MALWNN framework is evaluated through extensive simulation experiments. Subsequently, in Section VI, the actual performance of the models is evaluated using USRP and Raspberry Pi. We conclude the full paper in Section VII.
\section{RELATED WORKS}
% In recent years, a variety of lightweight AMC methods have emerged, including lightweight structure design, model compression and knowledge transfer, which are designed to enhance the efficiency and performance of AMC on edge devices. Besides, according to the number of samples, different AMC methods such as those based on supervised learning and weakly-supervised learning have also appeared. The following sections will conduct a detailed discussion on these AMC methods. 
In recent years, numerous lightweight AMC methodologies have been developed, encompassing technical directions such as lightweight architectural design, model compression, and knowledge distillation, with the objective of enhancing modulation recognition efficiency on resource-constrained edge devices. To address practical scenarios with limited training samples, novel AMC approaches based on weakly-supervised learning paradigms have been proposed. The following sections provide a systematic analysis of these technical approaches.
\subsection{Structural Design-based Methods}
% Methods based on structural design aim to reduce the demand for computational resources by devising more efficient network architectures\cite{KD},\cite{DWISE}. They strive to construct networks with fewer parameters or more efficient computing units. For instance, replacing standard convolution with grouped convolution can remarkably decrease the number of parameters and computational cost. Additionally, applying low-rank decomposition to perform a low-rank approximation on the weight matrix of the convolutional layer, which transforms a high-dimensional matrix into the product of multiple low-dimensional matrices, can significantly reduce the number of model parameters. The crux of this method lies in minimizing the network size and computational complexity while maintaining performance. The method based on Neural Architecture Search (NAS) can be regarded as a type of structural design approach. NAS is a method for automatically discovering efficient network architectures\cite{NAS1}\cite{NAS2}. It employs algorithms such as reinforcement learning or evolutionary algorithms to search for and optimize network structures. NAS aims to determine the optimal network architecture to balance key performance metrics such as accuracy, latency, and power consumption. 
Structural design methodologies for lightweight AMC focus on developing efficient network architectures to reduce computational demands \cite{KD, DWISE}. A prevalent strategy involves replacing standard convolutional layers with depthwise separable convolutions, significantly decreasing parameter counts while preserving feature extraction capabilities. Further complexity reduction can be achieved through low-rank decomposition techniques, which approximate convolutional weight matrices via products of lower-dimensional factors. Complementing these manual design approaches, Neural Architecture Search (NAS) employs reinforcement learning or evolutionary algorithms to automatically discover optimal network structures \cite{NAS1, NAS2}. This automated paradigm systematically explores architectural tradeoffs between classification accuracy and computational efficiency, particularly crucial for resource-constrained deployment scenarios.

% Based on the aforementioned concept, F. Zhang and his team designed a novel AMC network, PET-CGDNN, through the combination of modules like Parameter Estimation and Transformation (PET) and Gated Recurrent Unit (GRU)\cite{PETGCNN}. The performance of PET-CGDNN is approximate to that of MCLDNN, yet its model size is merely 20\% of the latter. L. Guo and his colleagues proposed ULCNN by integrating modules such as data augmentation, grouped convolution, and channel shuffling\cite{challenge}. The accuracy of ULCNN is on a par with existing models, with the number of parameters being less than 10,000. Moreover, following this concept, X. Wei et al.\cite{NAS1},\cite{NAS2} integrated NAS into AMC, leveraging continuous relaxation and gradient descent based on structural representations to optimize module connections. In contrast to traditional networks like ResNet, Inception, and MobileNet, the NAS-based AMC demonstrates enhanced recognition performance, reduced computational complexity, and a decreased number of model parameters. 
Recent advances in structural design have enabled the development of compact AMC networks through module recombination and NAS. Representative studies demonstrate 80\%-90\% parameter reduction while maintaining baseline accuracy by integrating specialized components (e.g., parameter estimation modules, grouped convolutions) or NAS-driven optimization \cite{PETGCNN, challenge, NAS1, NAS2}. Nevertheless, critical limitations persist: manual architectural engineering necessitates domain-specific expertise, NAS-based methods require intensive computational resources, and excessive compression frequently degrades recognition accuracy. To address these challenges, researchers are increasingly combining quantization techniques with knowledge distillation frameworks, enabling precision-preserving efficiency optimization through bit-width reduction and cross-model knowledge transfer.

\subsection{Methods Based on Model Compression and Knowledge Transfer }

Lightweight automatic modulation classification (AMC) methods primarily rely on pruning, quantization, and knowledge distillation. In pruning, Tu \textit{et al.}\cite{prune2} proposed an activation-guided pruning strategy to significantly compress the VT-CNN2 model while maintaining classification accuracy. Zhang \textit{et al.} \cite{PETGCNN} achieved parameter reduction in the PET-CGDNN model through sparse pruning without performance degradation. For quantization, researchers such as Tridgell \textit{et al.} \cite{quat1} implemented ternary weight quantization ($\{-1,0,+1\}$) for real-time inference on radio-frequency system-on-chip platforms, though full-precision models retained accuracy advantages. KD frameworks often involve transferring insights from complex architectures to lightweight counterparts, balancing computational efficiency and performance. Challenges persist in maintaining robustness under dynamic channel conditions and designing efficient distillation topologies \cite{prune2,PETGCNN,quat1}. However, quantization relies on hardware-specific calibration, pruning compromises structural integrity, and knowledge distillation is constrained by teacher-student architectural alignment.

\subsection{Weakly Supervised Learning-Based AMC }
Weakly-supervised learning aims to train models with limited annotation information, including semi-supervised learning and self-supervised learning methods. In the area of AMC under semi-supervised learning, Li \textit{et al.} \cite{Semi1} adopted a GAN-based method. A generator was used to generate fake samples, and a discriminator was used to judge the authenticity and modulation type of the samples. Dong \textit{et al}.\cite{Semi2} proposed the SSRCNN, which introduced the Kullback-Leibler (KL) divergence loss and cross-entropy loss for the unlabeled samples. However, when the number of labeled samples is small, it is difficult to provide reliable pseudo-labels for the unlabeled samples. Liu \textit{et al.} \cite{semi3} used self-supervised contrastive learning (CL) to pre-train the unlabeled samples, constructing positive pairs through rotation augmentation. However, these methods face challenges including error accumulation in pseudo-labels, modulation distortion from augmentation-induced time-frequency feature disruptions, high computational costs in contrastive frameworks, and static encoders' limited adaptability to dynamic environments.

As mentioned above, existing AMC methods face efficiency-accuracy tradeoffs akin to quantization/pruning/KD limitations, compounded by error accumulation in weakly supervised paradigms. Our ALWNN network and its few-shot variant MALWNN resolve these conflicts, delivering SOTA accuracy with ultra-low computation while excelling in generalization and operational efficiency\cite{BYOL,SIMLR,selfsemi1}.

\section{SIGNAL MODEL AND PROBLEM FORMULATION}
The modulation procedure relocates the spectrum of the baseband signal to a carrier signal of higher frequency. This facilitates the encoding and transmission of the baseband signal's information without disturbances. Moreover, it allows the signal to be adapted to the channel environment. A comprehensive depiction of the received signal can be articulated in this way:

\begin{equation}
r(t)=x(t)*c(t)+n(t),
\end{equation}

In the given formula, $x(t)$ stands for the transmitted signal in the absence of noise, $c(t)$ symbolizes the channel impulse response, and $n(t)$ is indicative of the noise, like Additive White Gaussian Noise (AWGN). The received signal $r(t)$ can be divided into its real and imaginary components, where the real component signifies the In-phase (I) channel and the imaginary component corresponds to the Quadrature (Q) channel. The issue of Deep Learning-based AMC can be described as:
\begin{equation}\hat{\mathrm{y}}=\arg\max_{\mathrm{y}\in\mathbf{Y}}f(\mathrm{y}|\mathbf{r};\mathbf{\theta}),\end{equation}

In this scenario, $y$ denotes the actual modulation type, and $\hat{y}$ stands for the predicted modulation type; $\mathbf{Y}$ is the pool of modulation types; ${f(\mathbf{\theta})}$ is the transformation function that maps samples to modulation types, where $\mathbf{\theta}$ symbolizes the model weight. In the sphere of Deep Learning and Automatic Modulation Classification, the main objective is to design a DL model, represented by $\mathbf{\theta}$, that strikes an optimal balance between high precision and low complexity. 

\section{ALWNN AND MALWNN}
In this section, the Adaptive Lightweight Wavelet Neural Network (ALWNN) is first introduced, and then the MALWNN framework is presented.
 
\subsection{Lightweight Adaptive Wavelet Neural Network }
The architecture of the ALWNN model is depicted in Figure 1. In the input phase, the raw data from the signal samples is converted into a tensor matrix of dimensions $N*1*2*L$, where $N$ is indicative of the batch size and $L$ represents the data length. The model first integrates I/Q channel data through depthwise separable convolutions, then extracts features via multi-level convolutional layers. An adaptive wavelet network performs multi-scale decomposition: iteratively refining low-frequency components while analyzing high-frequency components through global average pooling. Finally, hierarchical features are concatenated into a vector and processed by fully connected layers for classification.

\begin{figure*}[h]
  \centering
  \includegraphics[width=0.9\textwidth]{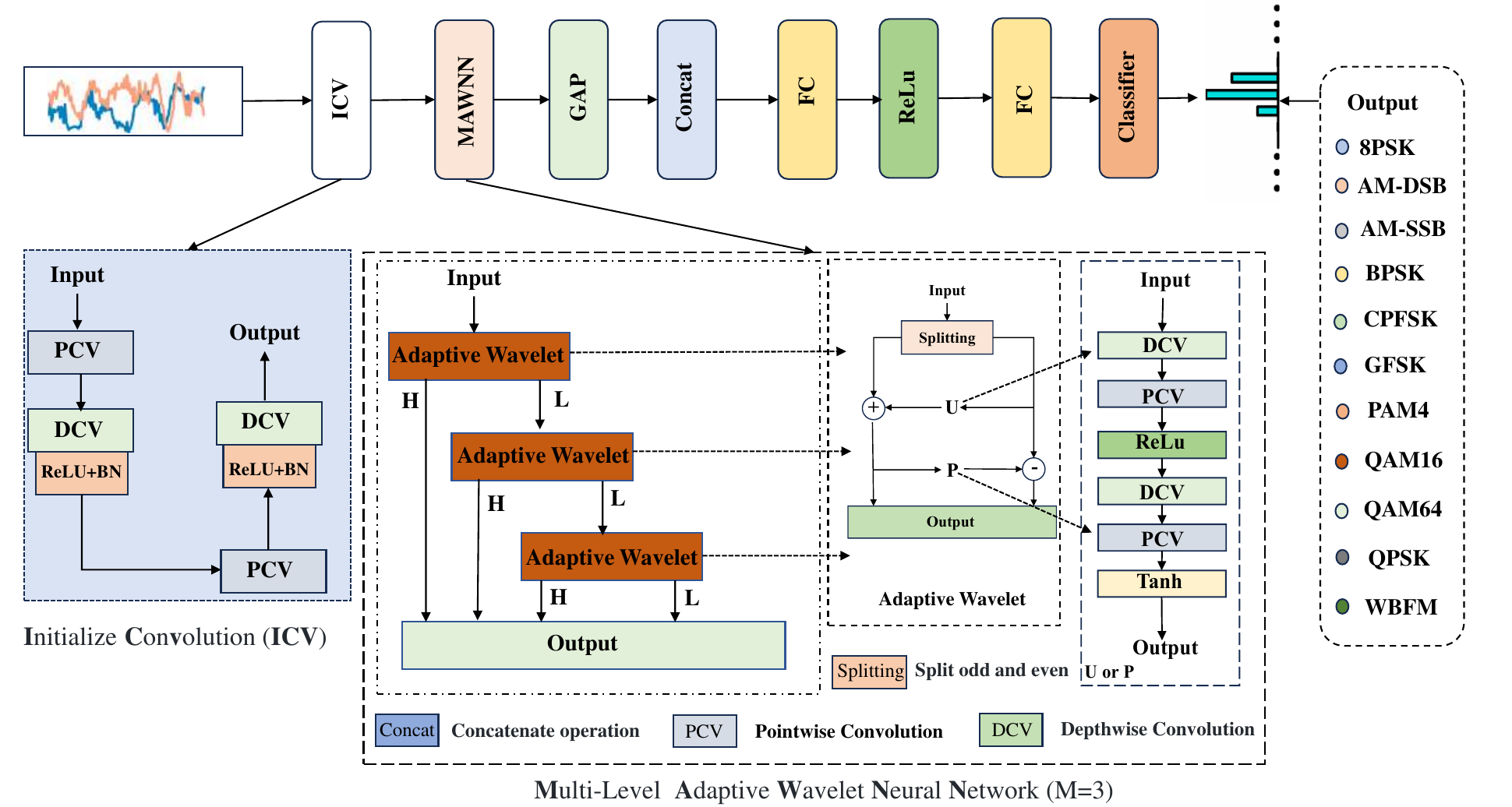}
  \vspace{-5pt}
  \caption{Architecture of our proposed ALWNN.}
  \vspace{-9pt}
\end{figure*}
\subsubsection{Initial Convolutions}
The signal employing I/Q modulation consists of two components, In-phase (I) and Quadrature (Q), which are orthogonal to each other. To encapsulate the interplay between the I channel and the Q channel, it's crucial to transform the tensor dimension signal from (2, L) to a tensor of dimension (1, L). This transformation was accomplished through the application of a deep convolutional layer and a pointwise convolutional layer. Within this deep convolutional layer, we utilized a total of 64 filters to amplify the variation of the features being extracted. In an effort to derive more advanced features, we applied an additional deep convolutional layer and a pointwise convolutional layer. As a result of this processing, the tensor's shape evolved from the initial state of (N, 1, 2, L) to a more complex form of (N, 64, L).
\subsubsection{Adaptive Wavelet}
The wavelet transform is a method that decomposes signals into components of different frequencies, enabling precise analysis at various scales by tweaking scale and shift parameters. It is particularly effective for transient signal analysis, offering insights into signal characteristics at multiple resolutions. Adaptive wavelets, an evolution of traditional wavelets, maintain core properties while introducing a flexible, dynamic approach to signal analysis. They adapt to the unique features of a signal, processing input $r$ to produce approximation $c$ and detail $d$ coefficients through a structured three-phase approach.

Splitting: The input signal $r$ is segmented into two distinct sections, with one comprising even-indexed components $r_{e}$ and the other odd-indexed components $r_{o}$. This division is achieved through a Split function, which categorizes the data into these even and odd segments based on their indices. It can be expressed as:
\begin{equation}
\begin{aligned}
\begin{aligned}\relax [r_o,r_e]=\text{Split}(r)\end{aligned} \\
r_o[k]=r[2k+1] \\
\begin{aligned}
r_e[k]=x[2k]
\end{aligned}
\end{aligned}
\end{equation}

Predictor: Leveraging the time-frequency relationship inherent in the signal, the Predictor $P(\cdot)$ is employed to estimate $r_{o}$ using $r_{e}$. The inherent correlation between $r_{o}$ and $r_{e}$ means that a well-chosen predictor $P(\cdot)$ can accurately forecast $r_{o}$, capturing the high-frequency details through the discrepancy between the actual and predicted values. This procedure can be articulated as:
\begin{equation}d=r_o-P(r_e)\end{equation}
where $d$ signifies high-frequency features, indicating the difference between $r_{o}$ and $P(r_e)$.

Updater: The Updater employs $U(\cdot)$ to refine $r_{e}$, integrating the high-frequency details $d$ into $U(\cdot)$ for enhancing the even components. This process of updating $r_{e}$ enables it to more accurately represent the low-frequency aspects of the original signal $r$. This mechanism can be illustrated as follows:
\begin{equation}c=r_e+U(d)\end{equation}

In classical wavelet transformations, functions $P(\cdot)$ and $U(\cdot)$ assume basic linear shapes. As an example, the elementary Haar wavelet is articulated as:
\begin{equation}\begin{aligned}d&=r_o-r_e\\c&=r_e+\frac{1}{2}d\end{aligned}\end{equation}

Nevertheless, conventional wavelet transformation techniques are not adaptable and don't fully utilize the benefits of data-driven deep learning. The static $P(\cdot)$ and $U(\cdot)$ functions could be substituted with neural networks for improved flexibility. This approach allows for the adaptive determination of wavelet coefficients, thereby overcoming the limitation of traditional wavelet bases which cannot be optimized through backpropagation.

\subsubsection{Multi-Level Adaptive Wavelet Neural Network}
Initial convolutional extraction ensures each channel in the feature map $\mathcal{F}$ encapsulates diverse feature information. Direct pooling of these feature vectors risks significant data loss. To mitigate this, adaptive wavelet transformation is applied using learnable $P(\cdot)$ and $U(\cdot)$. Consequently, $\mathcal{F}$ is decomposed into its low-frequency and high-frequency components, expressed as:
\begin{equation}\begin{aligned}
\left[L_o^{(j)},L_e^{(j)}\right]& =\text{Split}\left(L^{(j)}\right)  \\
H^{(j+1)}& =L_o^{(j)}-P{\left(L_e^{(j)}\right)}  \\
L^{(j+1)}& =L_e^{(j)}+U\left(H^{(j+1)}\right) 
\end{aligned}\end{equation}
In this context, $j$ indicates the iteration of adaptive wavelet transformations, with $H$ and $L$ representing high-frequency and low-frequency components, respectively. Notably, $L^{(0)}$ refers to the initial feature map derived from the first convolutional layer.

Figure 1 illustrates the architecture of $P(\cdot)$ and $U(\cdot)$. The process starts with reflection padding of the input low-frequency component to preserve sequence length post-convolution. This is followed by integration of feature map information via a depthwise convolution and a pointwise convolution layer, with ReLU activation function facilitating the polynomial fitting across different channels.

After conducting M levels of wavelet transformation, we obtain M high-frequency components ($H^{(1)}, H^{(2)}, ..., H^{(M)}$) and M low-frequency components ($L^{(1)}, L^{(2)}, ..., L^{(M)}$). Global average pooling is applied to these components, transforming them from $\mathbb{R}^{C\times L}$ to feature vectors $\mathcal{F}_{GAP} \in \mathbb{R}^{C\times1}$. Concatenating these vectors yields the final feature vector $X$, with $T$ denoting the vector length, as shown in:
\begin{equation}\begin{aligned}
\mathcal{F}_{GAP}(j)& =\frac1T\sum_{i=0}^{T-1}\mathcal{F}_{i,j},  \\
\text{X}& =\text{Concat}\Big(L_{GAP}^{(M)},H_{GAP}^{(1)},\ldots,H_{GAP}^{(M)}\Big).
\end{aligned}\end{equation}

\subsubsection{Loss Function}
The traditional wavelet scheme, constrained by predetermined formulas for predictors and updaters, couldn't optimally harness data. Alternatively, neural networks, renowned for their advanced learning abilities, improve this framework. The adaptive wavelet model is streamlined into two primary components: the detail loss function and the approximation loss function.
\begin{equation}\mathrm{Loss}_L = \sqrt{\sum_i (r_o[i] - P(r_e)[i])^2}\end{equation}
\begin{equation}\mathrm{Loss}_H = \sqrt{\sum_i (r_o[i] - r_e[i] - U(r_e)[i])^2}\end{equation}
The ALWNN training loss function integrates cross-entropy with $\operatorname{Loss}_L$ for approximation and $\operatorname{Loss}_H$ for detail reduction, optimizing data representation.
\begin{equation}
\begin{aligned}
L(\theta)&= \lambda_1\sum_{i=0}^{M-1}\left| {H}^{(i)} \right| + \lambda_2\sum_{i=0}^{M-2}\left\| {L}^{(i)} - {L}^{(i+1)} \right\|_2 \\
&- \log p{_\theta}(y=n|x)
\end{aligned}
\end{equation}

The batch training strategy is encapsulated in pseudocode detailed within Algorithm 1, where $\lambda_1$ and $\lambda_2$ serve to adjust the strength of regularization. $K$ specifies the variety of signal modulation categories, and $M$ indicates the depth of decomposition.
% \begin{algorithm}
% \caption{ALWNN Batch Training}
% \begin{algorithmic}[1]
% \REQUIRE Training data $r$ with targets $y$, model weights $w$, decomposition levels $M$, and regularization terms $\lambda_1$, $\lambda_2$.
% \STATE The initial convolutional layer processes the data and produces feature maps, denoted as $F$.
% \FOR{each $k \in [1, M]$}
%     \STATE $\text{Split}(F)$ divides $F$ into $F_{\text{even}}$ and $F_{\text{odd}}$ groups.
%     \STATE Predict: $H_k = F_{\text{odd}} - P_k(F_{\text{even}})$
%     \STATE Update: $L_k = F_{\text{even}} + U_k(H_i)$
%     \STATE Compute regularization terms: $\mathcal{L}_{H} += \lambda_1 \cdot \text{mean}(|H_k|)$; $\mathcal{L}_{L} += \lambda_2 \cdot |\text{mean}(L_k) - \text{mean}(F)|$
%     \STATE Concat feature vector: $f = \text{Concat}(f , \text{GAP}(H_k))$
%     \STATE Update $F$ with $L_k$: $F = L_k$
% \ENDFOR
% \STATE Feature fusion: $f = \text{Concat}(f , \text{GAP}(F))$.
% \STATE Compute $\hat{y} = \text{Softmax}(\text{FC}(f))$ and compute $\mathcal{L}_{CE}:-\sum_{j=1}^{K} y_j \log(\hat{y}_j)$
% \STATE Final loss is $\mathcal{L} = \mathcal{L}_{CE} + \mathcal{L}_{H} + \mathcal{L}_{L}$. Calculate the gradients during back-propagation $\frac{\partial \mathcal{L}}{\partial r}$ and update the model parameters $w = w - \eta \frac{\partial \mathcal{L}}{\partial w}$.
% \end{algorithmic}
% \end{algorithm}
\begin{algorithm}
\caption{ALWNN Batch Training}
\begin{algorithmic}[1]
\REQUIRE Training data $\mathbf{r}$ with targets $\mathbf{y}$, model weights $\mathbf{w}$, decomposition levels $M$, and regularization terms $\lambda_1$, $\lambda_2$.
\STATE The initial convolutional layer processes the data, resulting in feature maps, denoted as $\mathbf{F}$.
\FOR{each $k \in [1, M]$}
    \STATE $\text{Split}(\mathbf{F})$ partitions $\mathbf{F}$ into $\mathbf{F}_{\text{even}}$ and $\mathbf{F}_{\text{odd}}$.
    \STATE Compute prediction: $\mathbf{H}_k = \mathbf{F}_{\text{odd}} - P_k(\mathbf{F}_{\text{even}})$
    \STATE Update: $\mathbf{L}_k = \mathbf{F}_{\text{even}} + U_k(\mathbf{H}_i)$
    \STATE Calculate regularization terms: $\mathcal{L}_{H} = \mathcal{L}_{H} + \lambda_1 \cdot \text{mean}(|\mathbf{H}_k|)$; $\mathcal{L}_{L} = \mathcal{L}_{L} + \lambda_2 \cdot |\text{mean}(\mathbf{L}_k) - \text{mean}(\mathbf{F})|$
    \STATE Concatenate feature vector: $\mathbf{f} = [\mathbf{f}; \text{GAP}(\mathbf{H}_k)]$
    \STATE Update $\mathbf{F}$ to $\mathbf{L}_k$: $\mathbf{F} = \mathbf{L}_k$
\ENDFOR
\STATE Perform feature fusion: $\mathbf{f} = [\mathbf{f}; \text{GAP}(\mathbf{F})]$.
\STATE calculate $\mathcal{L}_{CE}:-\sum_{j=1}^{K} y_j \log(\hat{y}_j)$
\STATE The final loss is $\mathcal{L} = \mathcal{L}_{CE} + \mathcal{L}_{H} + \mathcal{L}_{L}$. Compute the gradients $\frac{\partial \mathcal{L}}{\partial \mathbf{r}}$ during back-propagation and update the model parameters: $\mathbf{\theta} = \mathbf{\theta} - \eta \frac{\partial \mathcal{L}}{\partial \mathbf{\theta}}$.
\end{algorithmic}
\end{algorithm}
\subsection{Few-Shot AMC Framework}
\begin{figure*}[h]
  \centering
  \includegraphics[width=0.9\textwidth]{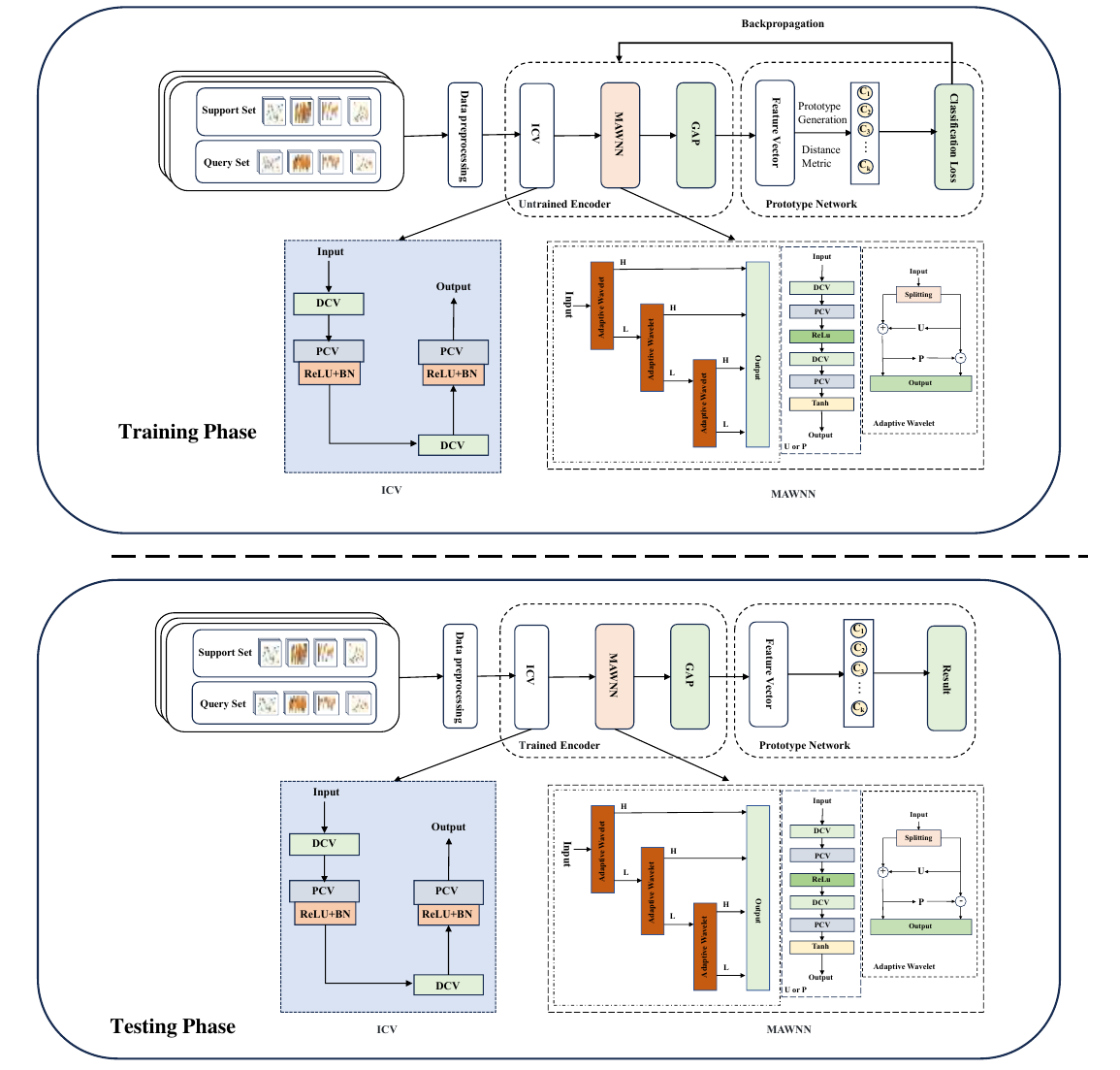}
  \vspace{-5pt}
  \caption{Architecture of our proposed few shot framework MALWNN.}
  \vspace{-9pt}
\end{figure*}

Figure 2 presents the framework structure of MALWNN. Our system is comprised of a few-shot training module and a few-shot testing module. The training module contains a data processing section, a feature extraction portion, and a class prototype component. The data processing part primarily extends or truncates the signals to guarantee a signal length of 1024. The feature extraction module herein is the ALWNN referred to previously, which is capable of fully extracting the features of various dimensions of the signal while reducing the parameter count and computational complexity of the feature extraction network. The signal samples are mapped into a unified feature measurement space via the feature extraction module, and the class prototype module conducts distance measurement and ascertains the modulation pattern of the signal to be identified. The Euclidean distance is employed as the distance measurement approach in this paper. Owing to the attainment of meta-knowledge, this method can sustain favorable performance even with an insufficient number of samples. The fundamental structure of the testing module resembles that of the training module, with the exception that no backpropagation of gradients is performed. Notably, the testing module directly yields the classification outcome, whereas the training result outputs the loss function value. 
\subsubsection{Meta Training}
The training module conducts the training of the parameter $\theta$ of ALWNN. In this stage, the training is carried out episodically. Each episode, marked as $\epsilon$, consists of a support set for prototype generation and a query set for modulation prediction and parameter update. To generate the support set and query set for each episode, we first randomly select $n$ categories from the source dataset. Then, within each selected category, we further randomly pick $k$ instances. Here, $n$ represents the total number of classes within the support set, commonly referred to as $n$-way, and $k$ represents the number of data samples for each class (way), known as $k$-shot. The total number of episodes, denoted as $N_{\epsilon}$, can be determined by the following formula:

\begin{equation}
N_{\epsilon}=\frac{p_{train} \cdot N}{N_{S}+N_{Q}} \cdot N_{epoch}
\end{equation}

Among them, $N$ represents the total quantity of data. $p_{train}$ refers to the proportion of the training dataset. $N_{S}$ stands for the number of support sets, and $N_{Q}$ represents the number of query sets. $N_{epoch}$ denotes the number of training epochs, The $N$ annotated data used as input, denoted as $D = \{(x_1, y_1), \ldots, (x_N, y_N)\}$. In each episode, the preprocessing module guarantees a fixed signal length by means of signal truncation and expansion. Thereafter, the data of the support set and the query set are divided into small segments as previously depicted and then input into the encoder of the ALWNN network. For the support set, the prototype $p_l$ is generated by averaging the feature vectors extracted from the annotated dataset $S_l$ that belongs to class $l$, as shown in the following equation. 

\begin{equation}p_l=\frac{1}{|D_l|}\sum_{(x_j,y_j)\in D_l}f_\theta(x_j)\end{equation}

The feature vectors extracted from the query set are classified using the generated prototypes based on a distance function $d$, which can be methods such as the Euclidean distance or cosine similarity. In this paper, the Euclidean distance function is used as the distance function. Based on the softmax function of the distance between the query point $x$ and the prototypes in the embedding space, we generate a distribution over classes. The formula for this distribution is as follows:
\begin{equation}p_\theta(y=l|x)=\frac{\exp(-d(f_\theta(x),p_l))}{\sum_{l^{\prime}}\exp(-d(f_\theta(x),p_{l^{\prime}}))}\end{equation}
Within each training episode, the parameter $\theta$ is iteratively updated using the Adam optimizer to minimize the loss function. The specific training algorithm can be found in Algorithm 2. The function is expressed as:

\begin{equation}
\begin{aligned}
L(\theta)&= - \log p{_\theta}(y=l|x) + \lambda_1\sum_{i=0}^{M-1}\left| {H}^{(i)} \right| \\
&+ \lambda_2\sum_{i=0}^{M-2}\left\| {L}^{(i)} - {L}^{(i+1)} \right\|_2 
\end{aligned}
\end{equation}

\begin{algorithm}[H]
\caption{Meta Training Workflow}
\begin{algorithmic}[1]
\STATE {$n$ Is the Number of Classes per Episode, $E$ Is the Selected $n$ Classes for Episode, $N_S$ Is the Number of Support samples per Class, $N_Q$ Is the Number of Query samples per Class, $\hat{m}$ Is the Bias-Corrected Moving Average of the Gradients, $\hat{v}$ Is the Bias-Corrected Moving Average of the Squared Gradients, $\alpha$ Is the Learning Rate and $\epsilon$ Is a Small Value Used for Numerical Stability. Random uniform$(D,N)$ Denotes Uniform and Random Selection of $N$ Values From the $D$ Set. Signal Length$(D,z)$ Denotes the Adjustment of All $x$ Lengths in Set $D$ to $z$.}
\STATE {\bfseries Input:} Training set $D_{train} = \{(x_1, y_1), \ldots,(x_N, y_N)\}$
\STATE {\bfseries Output:} Trained ALWNN $f_{\theta}$
\FOR {$l = 1,\ldots,n$}
\STATE $D_{support} \leftarrow \text{Random uniform}(D_{E_l},N_S)$
\STATE $D_{query} \leftarrow \text{Random uniform}(D_{E_l}\backslash D_{support},N_Q)$
\STATE $D_{support} \leftarrow \text{Signal Length}(D_{support},z_m)$
\STATE $D_{query} \leftarrow \text{Signal Length}(D_{query},z_m)$
\STATE $p_l \leftarrow \frac{1}{|D_l|}\sum_{(x_j,y_j)\in D_l}f_{\theta}(x_j)$
\ENDFOR
\STATE $L \leftarrow 0$ {Initialize loss $L$}
\FOR {$l = 1,\ldots,n$}
\FOR {$(x, y) \in D_{query}$}
\STATE $\theta \leftarrow \theta - \frac{\alpha}{\sqrt{\hat{v}+\epsilon}}\hat{m}$
\ENDFOR
\ENDFOR
\end{algorithmic}
\end{algorithm}
\subsubsection{Meta Testing}
The testing module utilizes the model weights trained through training truncation, which remain fixed during the testing process. In the meta-testing stage, both the support set and the query set contain unseen modulation methods, enabling us to evaluate the model's adaptability to new domains and its generalization ability. The signal length processed by the data processing module is 1024, which is the signal length of RadioML2018.01a. In most few-shot learning (FSL)-based methods, the commonly adopted configuration for the support set is the 5-shot setting, meaning that the support set contains five data samples. The testing module uses the trained ALWNN model to generate $n^{\prime}$ prototypes, where $n^{\prime}$ represents the number of target classes for testing. The query set used for inference is classified based on the Euclidean distance between the embedding vectors and the prototypes.

\section{NUMERICAL RESULTS}
\subsection{Datasets}
The experiment was carried out using the RML2016.10a, RML2016.10b, and RML2018.01a datasets, which were synthesized via GNU Radio to accurately mimic the wireless conditions in the real world, incorporating elements like multipath fading, sampling rate offset, additive white noise, and the central frequency offset of the wireless channel. The RML2016.10a dataset contains 11 prevalent modulation types, totaling 220,000 modulated signals, including 8PSK, QPSK, BPSK, QAM64, QAM16, CPFSK, GFSK, 4PAM, WBFM, AM - SSB, and AM - DSB. The RML2016.10b dataset, on the other hand, holds 12 million signals across 10 types, excluding AM - SSB, and in both these datasets, the SNR ranges from -20dB to 18dB in 2dB steps, with each signal stored as a 2×128 matrix representing the in-phase and quadrature parts of the modulated signal samples. The RML2018.01a dataset is larger and more complex, featuring 24 modulation formats and 255,590,400 modulated signals, with an SNR from -20dB to 30dB at 2dB intervals and a signal size of 2×1024. It simulates more channel impairments, increasing classification difficulty, and also incorporates more high-order and analog modulation formats, making classification highly challenging.
\subsection{Experimental Settings}
% In the experiment of lightweight ALWNN, the datasets were split, with 60\% allocated for training, 20\% for validation, and the remaining 20\% used as a test set. These divisions were utilized to train and evaluate all tasks. Stratified sampling was performed for different signal-to-noise ratios of different modulation formats. Each signal-to-noise ratio of different format signals was also divided in a 6:2:2 ratio. The experiment was implemented based on the Pytorch 2.0 framework. In the experiments on RML2016.10a and RML2016.10b, we used an Adam optimizer with a batch size of 256 to train the proposed network. The initial learning rate was established at 0.001 for the experiment. The regularization term coefficients $\lambda_{1}$ and $\lambda_{2}$ are equal to 0.01. Should the validation set loss fail to decrease over a span of five epochs, an early termination of the training process will be invoked. The number of layers of the adaptive wavelet transform is set to 1, because the signal length of the data is 128, and using a 1-layer adaptive wavelet transform is sufficient to extract features. The depth convolutional kernels for initializing the convolutional layers are of sizes 2 × 7 and 1 × 5 respectively. The depth convolutional kernels for the $U(\cdot)$ and $P(\cdot)$ parts are of size 1 × 5. The kernels for pointwise convolutions throughout the model are of size 1 × 1. All experiments were conducted using NVIDIA CUDA with a GeForce RTX 4090 GPU.

In the experiment of lightweight ALWNN, the datasets were partitioned, with 60\% being allocated for training, 20\% for validation, and the remaining 20\% serving as the test set. These partitions were employed to train and evaluate all tasks. Stratified sampling was carried out for different signal-to-noise ratios of various modulation formats. Each signal-to-noise ratio of different format signals was also divided in a ratio of 6:2:2. This experiment was implemented based on the Pytorch 2.0 framework. In the experiments concerning RML2016.10a and RML2016.10b, we utilized an Adam optimizer with a batch size of 256 to train the proposed network. The initial learning rate was set at 0.001 for this experiment. The regularization term coefficients $\lambda_1$ and $\lambda_2$ were equal to 0.01. If the loss of the validation set failed to decrease within a span of five epochs, an early termination of the training process would be triggered. For the RML2016.10A and RML2016.10b datasets, the number of layers of the adaptive wavelet transform was set to 1. In contrast, for the RML2018.01a dataset, the number of layers of the adaptive wavelet transform was set to 3. This is because when the signal length of the data is 128, a single layer of the adaptive wavelet transform is sufficient to extract features, while a signal length of 1024 renders a single layer far from adequate. The sizes of the depth convolutional kernels for initializing the convolutional layers were $2\times7$ and $1\times5$ respectively. The depth convolutional kernels for the $U(\cdot)$ and $P(\cdot)$ parts were of size $1\times5$. The kernels for pointwise convolutions throughout the model were of size $1\times1$. All experiments were conducted using NVIDIA CUDA with a GeForce RTX 4090 GPU.

In the experiment of MALWNN, we selected the RML2018.10a dataset and divided it proportionally. Among it, 70\% of the data was used for training the model, 10\% of the data served as the validation set for optimizing the model parameters, and the remaining 20\% of the data was used as the test set to evaluate the model performance. 
This experiment was implemented based on the Pytorch 2.0 framework. During the training process, we used an Adam optimizer with a batch size set to 512 to train and optimize the proposed network. Meanwhile, the initial learning rate of the experiment was set to 0.001, and the regularization term coefficients $\lambda_1$ and $\lambda_2$ were both set to 0.001. In view of the characteristics of the dataset, the number of layers of the adaptive wavelet transform was determined to be 3, and the Euclidean distance metric was adopted as the distance measurement method. The other relevant settings remained the same as those mentioned above.

\subsection{Classification Performance of ALWNN}

\begin{table*}[h]
\centering
\caption{Combined Performance of All Methods on Different Datasets}
\begin{tabular}{lcccccccc}
\toprule
Dataset & Model & Accuracy & MF1 & Kappa & Parameters(K) & MACC(M) & FLOPS(M) &Inference time (ms/sample) \\
\midrule
\multirow{7}{*}{RML2016.10a}
& ALWNN & \textbf{\color{blue} 0.6214} & \textbf{\color{blue} 0.6423} & \textbf{\color{blue} 0.5815} & \textbf{\color{red} 9.899K} & \textbf{\color{red} 0.53M} & \textbf{\color{red} 0.52M} & \textbf{\color{red} 0.045} \\
& CLDNN & 0.5871 & 0.6091 & 0.5447 & 160K & 5.74M & 11.986M & 0.225 \\
& MCLDNN & 0.6203 & 0.6381 & 0.5803 & 380K & 20.94M & 41.876M & 0.283\\
& ICAMC & 0.5681 & 0.5983 & 0.5401 & 1210K & 7.7M & 15.403M & 0.102\\
& AMC - Net & \textbf{\color{red} 0.6251} & \textbf{\color{red} 0.6483} & \textbf{\color{red} 0.5885} & 470K & 9.4M & 18.799M & 0.122\\
& CDSCNN & 0.5920 & 0.6060 & 0.5512 & 860K & 13M & 25.998M & 0.324\\
& MCNET & 0.5600 & 0.5912 & 0.5382 & \textbf{\color{blue} 120K} & \textbf{\color{blue} 3.44M} & \textbf{\color{blue} 9.289M} & \textbf{\color{blue} 0.088} \\
\midrule
\multirow{7}{*}{RML2016.10b}
& ALWNN & \textbf{\color{blue} 0.6393} & \textbf{\color{blue} 0.6386} & \textbf{\color{blue} 0.5992} & \textbf{\color{red} 9.7K} & \textbf{\color{red} 0.53M} & \textbf{\color{red} 0.52M} & \textbf{\color{red} 0.047} \\
& CLDNN & 0.6021 & 0.6088 & 0.5972 & 160K & 5.74M & 11.986M & 0.216 \\
& MCLDNN & 0.6358 & 0.6412 & 0.6028 & 380K & 20.94M & 41.876M & 0.282 \\
& ICAMC & 0.6243 & 0.6332 & 0.6023 & 1210K & 7.7M & 15.403M & 0.106\\
& AMC - Net & \textbf{\color{red} 0.6463} & \textbf{\color{red} 0.6487} & \textbf{\color{red} 0.6081} & 470K & 9.4M & 18.799M & 0.132 \\
& CDSCNN & 0.6288 & 0.626 & 0.587 & 860K & 13M & 25.998M & 0.323\\
& MCNET & 0.6074 & 0.6102 & 0.5883 & 120K & \textbf{\color{blue} 3.44M} & \textbf{\color{blue} 9.289M} & \textbf{\color{blue} 0.082} \\
\midrule
\multirow{7}{*}{RML2018.01a}
& ALWNN & \textbf{\color{blue} 0.6259} & \textbf{\color{blue} 0.6266} & \textbf{\color{blue} 0.6097} & \textbf{\color{red} 31K} & \textbf{\color{red} 2.12M} &\textbf{\color{red} 2.132M} & \textbf{\color{red} 0.090}\\
& CLDNN & 0.5549 & 0.5602 & 0.5432 & 0.84M & 30.135M & 68.916M & 1.431\\
& MCLDNN & 0.6242 & 0.6222 & 0.6012 & 0.38M & 210.94M & 241.876M & 1.963 \\
& ICAMC & 0.5839 & 0.5913 & 0.5723 & 8.21M & 53.9M & 93.85M & 0.125\\
& AMC - Net & \textbf{\color{red} 0.6263} & \textbf{\color{red} 0.6287} & \textbf{\color{red} 0.611} & 28.2M & 56.4M & 197.88M & 1.231\\
& CDSCNN & 0.6088 & 0.6166 & 0.57 & 56M & 121.3M & 205.8M & 2.032 \\
& MCNET & 0.5619 & 0.5645 & 0.5422 & \textbf{\color{blue} 120K} & \textbf{\color{blue} 7.44M} & \textbf{\color{blue} 11.9M} & \textbf{\color{blue} 0.108} \\
\bottomrule
\end{tabular}
\end{table*}

We compared our ALWNN model with six benchmark models using the RadioML2016.10a and RadioML2016.10b datasets. The primary benchmark models include CLDNN\cite{CLDNN}, MCLDNN\cite{MCLDNN}, ICAMC\cite{ICAMC}, MCNET\cite{MCNET}, AMC-Net\cite{AMC}, CDSCNN\cite{CDSCNN}. All methods employed the original I/Q samples as input.

As shown in Table I, we utilized multiple key metrics to conduct a detailed and comprehensive evaluation and comparison analysis of our self-developed Adaptive Wavelet Neural Network (ALWNN) model against other benchmark models on the RadioML2016.10a and RadioML2016.10b datasets. The selected metrics cover important dimensions such as accuracy, kappa coefficient, Macro - F1 score (MF1), number of parameters, floating - point operations per second (FLOPS), and normalized multiply - accumulate complexity (NMACC).

Among them, the accuracy precisely reflects the average recognition accuracy level of the model under all SNR conditions and is one of the crucial fundamental metrics for evaluating the performance of the model. The kappa coefficient, as a statistical measure specifically used to evaluate the degree of classification consistency, in this context, its core function lies in accurately assessing the degree of fit and consistency between the model's classification results and the actual true annotation results. The Macro - F1 score (MF1) approaches from the perspective of comprehensive evaluation across multiple categories. By averaging the F1 values of each category, it enables us to comprehensively and evenly grasp the overall performance status of the model across multiple categories, effectively avoiding the problem of overlooking the overall performance balance due to excessive focus on some categories, thus providing a more comprehensive, objective, and accurate perspective for the comprehensive performance evaluation of the model. The number of parameters, to a certain extent, directly reflects the complexity of the model's structure and its information - carrying capacity, indirectly reflecting the scale and capacity characteristics of the model. The floating - point operations per second (FLOPS) emphasizes characterizing the level of computational efficiency of the model during operation, directly related to the actual demand for computational resources and the speed of resource consumption when the model is running. The normalized multiply - accumulate complexity (NMACC) further clearly reflects the complex characteristics and resource demand tendencies of the model from a specific computational complexity dimension.

Especially noteworthy is that the accuracy performance of our developed Adaptive Wavelet Neural Network model is only slightly inferior to that of the optimal AMCNet model, demonstrating strong competitiveness. At the same time, compared with the baseline methods, the ALWNN model has achieved remarkable results in terms of computational efficiency and resource demand optimization. Specifically, its floating - point operations per second have been significantly reduced by 1.25 to 1.91 orders of magnitude, and the normalized multiply - accumulate complexity has also been significantly reduced by 0.81 to 1.6 orders of magnitude. This dual reduction in computational volume and complexity is expected to sharply reduce the inference latency of the model on low - computational - power central processing units (such as Raspberry Pi or microcontrollers) from the original several seconds to the millisecond level. At the same time, it can also effectively alleviate the storage pressure and computational resource demand burden on terminal devices. These series of advantages make the practical application of the adaptive modulation and coding method in wireless communication systems possible, laying a solid foundation for its wide promotion and application, and is expected to promote the further development and innovation in the field of wireless communication technology. 
Figure 3, Figure 4 and Figure 5 clearly present the comparison of the classification performance between our self-developed Adaptive Wavelet Neural Network (ALWNN) model and other benchmark models under different SNR situations on the RadioML2016.10a, RadioML2016.10b and RadioML2018.01a datasets. It can be clearly seen from the presented data that under numerous different SNR conditions, our ALWNN model demonstrates outstanding performance advantages and its performance surpasses that of most of the models it is compared with, which fully proves the high efficiency and reliability of this model in complex signal classification tasks. 
\subsection{Ablation Study}
The results of the ablation experiment are presented in Table II and Table III. In Table II, a study was carried out on the impact of each module of ALWNN on the model performance. It can be clearly observed from Table II that the AWN module has an extremely significant impact on the model performance. This is due to its ability to extract features from multiple dimensions, thus enhancing the model's understanding and processing ability of the data. Next is the initialization convolution module, which also plays a crucial role in improving the model performance. Furthermore, the introduction of the attention mechanism and the increase in the number of AWN stages can also improve the model performance, but this will correspondingly lead to an increase in the number of parameters.
The experiment in Table III shows that for the RML2018.01a dataset, the model can achieve the best performance when M = 3. When M = 0, the recognition accuracy is at the lowest level, and the model has difficulty in accurately classifying the data. After reaching the optimal state at M = 3, as the value of M further increases, the accuracy of the model generally remains stable and no obvious upward trend appears. At the same time, the number of parameters increases significantly. We infer that the reason for this phenomenon is that the signal length of RML2018.01a is 1024. When the number of decomposition layers is low, the model cannot efficiently extract multi-dimensional features. When the number of decomposition stages exceeds a certain threshold, the data features have been fully extracted, so it is difficult to further improve the classification performance. 

\vspace{-9pt}
\begin{figure}[h]
  \centering
  \includegraphics[width=0.5\textwidth]{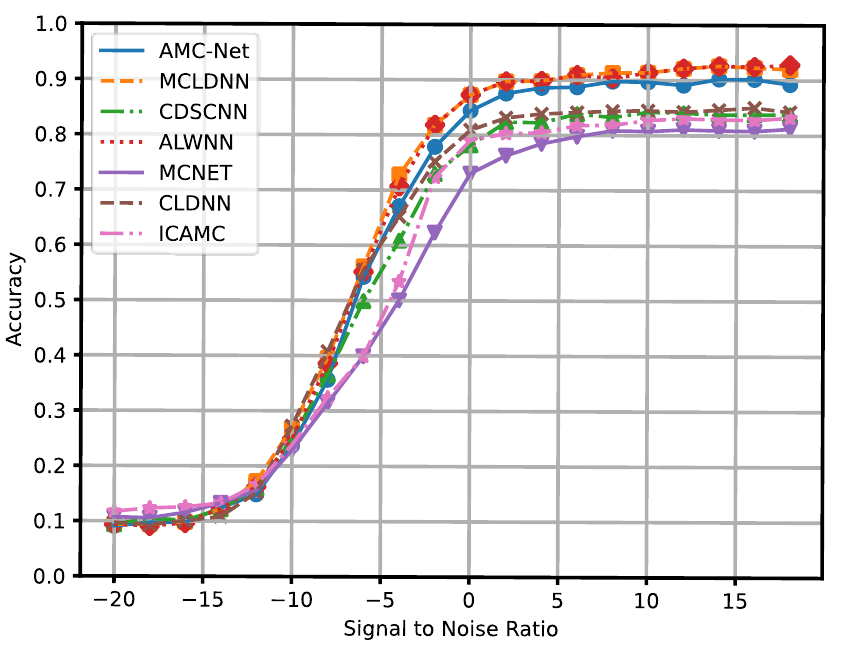}
  \vspace{-20pt}
  \caption{Classification accuracy on different SNR of ALWNN and other models on RML2016.10a.}
  \vspace{-9pt}

\end{figure}
\vspace{-9pt}
\begin{figure}[h]
  \centering
  \includegraphics[width=0.5\textwidth]{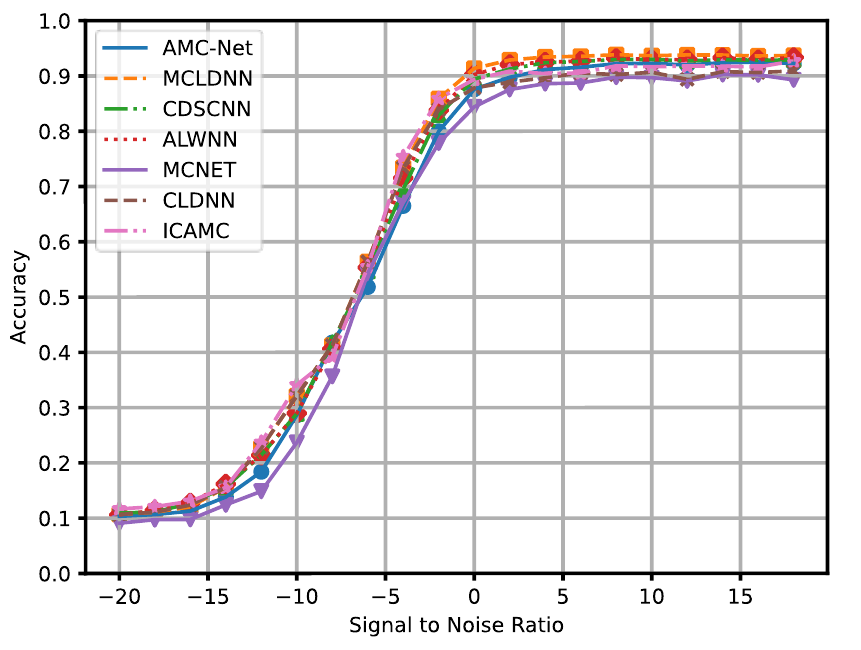}
  \vspace{-20pt}
  \caption{Classification accuracy on different SNR of ALWNN and other models on RML2016.10b.}
  \vspace{-9pt}
\end{figure}
\vspace{-9pt}
\begin{figure}[h]
  \centering
  \includegraphics[width=0.5\textwidth]{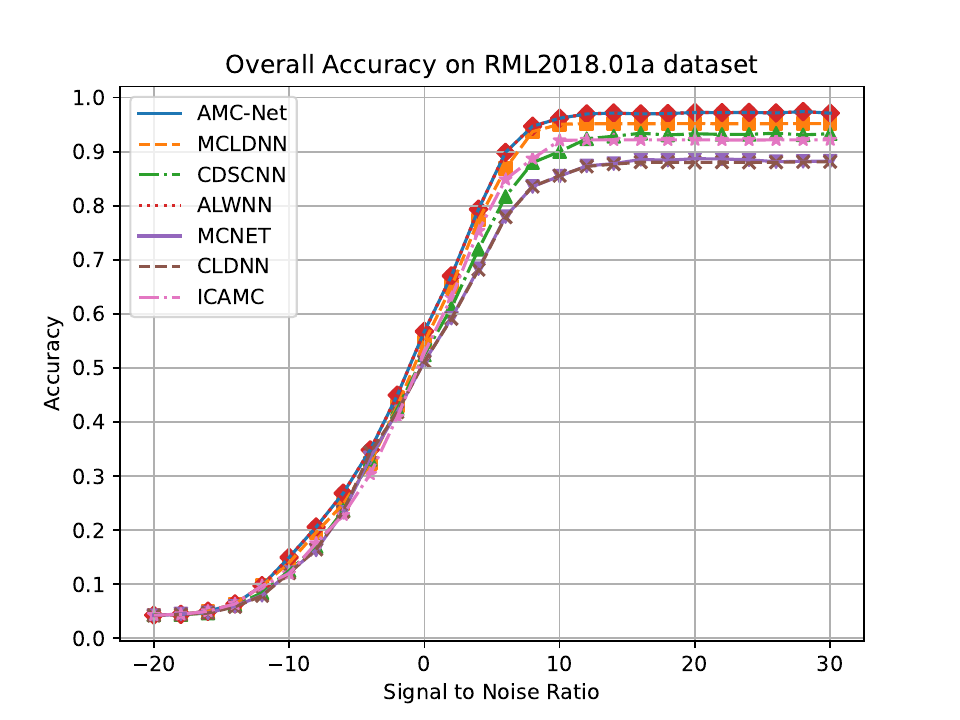}
  \vspace{-10pt}
  \caption{Classification accuracy on different SNR of ALWNN and other models on RML2018.01a.}
  \vspace{-9pt}
\end{figure}

\begin{table*}[h]
\centering
\caption{THE ABLATION STUDIES}
\begin{tabular}{lcccccc}
\hline
Method & Ablation part & Average accuracy & MF1 & kappa & Number of Parameters(M) \\
\hline
ALWNN & - & $62.59\%$ & $62.66\%$ & $60.97\%$ & $0.031$ \\
ALWNN - A & add attention & $62.60\%$ ($0.015\%\uparrow$) & $62.94\%$ ($0.44\%\uparrow$) & $60.93\%$ ($0.14\%\downarrow$) & $0.08$ ($110\%\uparrow$) \\
ALWNN - B & w/o CV & $60.56\%$ ($3.240\%\downarrow$) & $60.36\%$ ($3.67\%\downarrow$) & $59.60\%$ ($2.24\%\downarrow$) & $0.025$ ($0.6\%\downarrow$)\\
ALWNN - C & w/o AWN & $54.16\%$ ($13.53\%\downarrow$) & $54.45\%$ ($13.10\%\downarrow$) & $53.12\%$ ($12.87\%\downarrow$) & $0.015$ ($51.61\%\downarrow$) \\
ALWNN - D & ADD Harr & $58.39\%$ ($6.71\%\downarrow$) & $59.13\%$ ($5.63\%\downarrow$) & $57.23\%$ ($6.13\%\downarrow$) & $0.029$ ($6.45\%\downarrow$) \\
ALWNN - E & add FFT & $60.99\%$ ($2.55\%\downarrow$) & $60.73\%$ ($3.08\%\downarrow$) & $58.59\%$ ($3.90\%\downarrow$) & $0.032$ ($3.22\%\uparrow$) \\
ALWNN - F & M=1 & $60.50\%$ ($3.33\%\downarrow$) & $60.30\%$ ($3.76\%\downarrow$) & $58.89\%$ ($3.41\%\downarrow$) & $0.02336$ ($24.64\%\downarrow$) \\
\hline
\end{tabular}
\end{table*}
\begin{table}[h]
\centering
\caption{Sensitivity Analysis of Parameter M}
\begin{tabular}{lccc}
\hline
Num of Level & Average accuracy & MF1 & Number of Parameters(K) \\
\hline
$M = 0$ & $54.16\%$ & $54.45\%$ & $15.19$ \\
$M = 1$  & $58.50\%$  & $59.30\%$  & $23.36$  \\
$M = 2$  & $61.59\%$  & $61.82\%$  & $35.39$  \\
$M = 3$  & $62.59\%$  & $62.66\%$  & $57.08$  \\
$M = 4$ & $62.60\%$  & $62.66\%$  & $81.54$  \\
$M = 5$ & $62.61\%$  & $62.63\%$  & $129.42$  \\
\hline
\end{tabular}
\end{table}

\vspace{20pt}
\subsection{Few Shot Framework Evaluation}
\begin{table*}[h]
\centering
\caption{Experimental data set diversity}
\begin{tabular}{>{\centering\arraybackslash}m{1cm}ccc}
\toprule
case & Train set & Test set \\
\midrule
A & 4ASK, 8ASK, QPSK, 8PSK, 32PSK,  \\
  & 32APSK, 128APSK, 32QAM, 64QAM, 256QAM & OOK, BPSK, 16APSK, 16QAM, GMSK \\
\midrule
B & OOK, 4ASK, BPSK, 8PSK, 16PSK,  \\
  & 32PSK, 64APSK, 64QAM, 16QAM,128APSK 
  & AM - DSB - SC, AM - SSB - WC, AM - DSB - WC, FM, AM - SSB - SC \\
\midrule
C & OOK, 4ASK, BPSK, 16PSK, 8PSK,  \\
  & 16APSK, 32APSK, 128QAM, 64QAM, 32QAM & 16QAM, QPSK, 8PSK, 16PSK, 32PSK \\
\midrule
D & OOK, 8ASK, 16PSK, 16APSK, 32APSK,  \\
  & AM - SSB - SC, AM - DSB - WC, 128APSK, 32QAM, GMSK & 4ASK, 8PSK, FM, 256QAM, 32APSK \\
\midrule
E & ALL & ALL \\
\bottomrule
\end{tabular}
\end{table*}
% In this part, we evaluated the adaptive performance of our proposed Meta-ALWNN framework to new modulation types. As mentioned before, one of the advantages of meta-learning is that it can quickly adjust the model to adapt to new and unknown classes. For example, consider a scenario where a software-defined radio (SDR) is in operation and the device needs to be upgraded to recognize new modulation types. We conducted an experiment in which the proposed model was trained on 10 randomly selected modulation methods out of a total of 24 modulations. Then, we randomly selected 5 modulations from the remaining 12 modulations for testing. We divided the test cases into 5 categories as shown in Table IV. For each test case, we performed 100 test iterations, and each iteration involved the random selection of 5 test modulation data. Subsequently, we calculated the average precision. The default value of the number of samples (shot) was set to 5. The reason is that many few-shot learning studies use 1-shot and 5-shot evaluations as benchmark references. The number of sample frames used for training was approximately 1.3 million frames, while the number of sample frames used for testing was about 500,000 frames.

This experiment evaluates the adaptive performance of the proposed MALWNN framework to new modulation types. Leveraging the rapid transfer capability of meta-learning (e.g., when software-defined radios require dynamic recognition of new modulations), we established a test scenario: 10 modulation types were randomly selected from 24 candidates for model training, while 5 unseen types were sampled from the remaining 12 for testing (see Table IV). Adopting the 5-shot configuration aligned with few-shot learning benchmarks, each test group underwent 100 random trials with averaged precision calculation. The training and testing datasets contained approximately 1.3 million and 500,000 sample frames respectively.

Figure 6 depicts the precision results of three test cases, illustrating that in the testing phase, the average precision of the five modulation methods in the high signal-to-noise ratio region was approximately 84\%. The variation in accuracy among different test cases was affected by the modulation methods used in the training phase. When similar modulation methods appeared in the training phase, the model performance often tended to be better in the testing phase, such as case D and case E. In case B, all the modulation methods in the training phase were digital modulations, while all those in the testing phase were analog modulations, so the testing performance was relatively poor.

Figure 7 presents the experimental results of investigating the impact of the Shot value on the model performance in the testing phase. For the five-class classification problem in the testing phase, the Shot values were 1, 5, 10, 15, and 20 respectively. The results showed that the accuracy increased with the increase of the Shot value. When the Shot value was greater than or equal to 15, our method achieved an accuracy rate of 95\%, demonstrating that the proposed framework can achieve good performance even with only a small number of datasets.

In this section, to verify the performance superiority of the algorithm proposed in this paper compared with other few-shot modulation recognition algorithms, we selected few-shot modulation recognition algorithms based on data augmentation (DA) and transfer learning (TL) for comparison experiments. Additionally, three other meta-learning algorithms were chosen, namely the relation network (RN), matching networks (MN), and model-agnostic meta learning (MAML).

\vspace{-9pt}
\begin{figure}[h]
  \centering
  \includegraphics[width=0.5\textwidth]{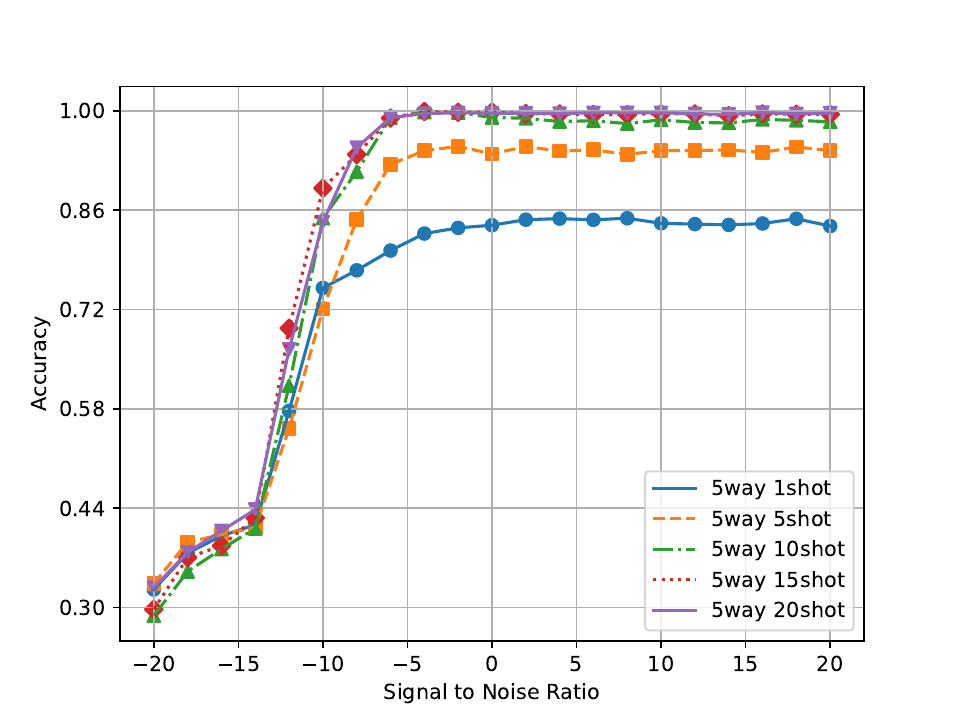}
  \vspace{-10pt}
  \caption{Classification accuracy on different shot of on Case A.}
  \vspace{-9pt}
\end{figure}

\vspace{-9pt}
\begin{figure}[h]
  \centering
  \includegraphics[width=0.5\textwidth]{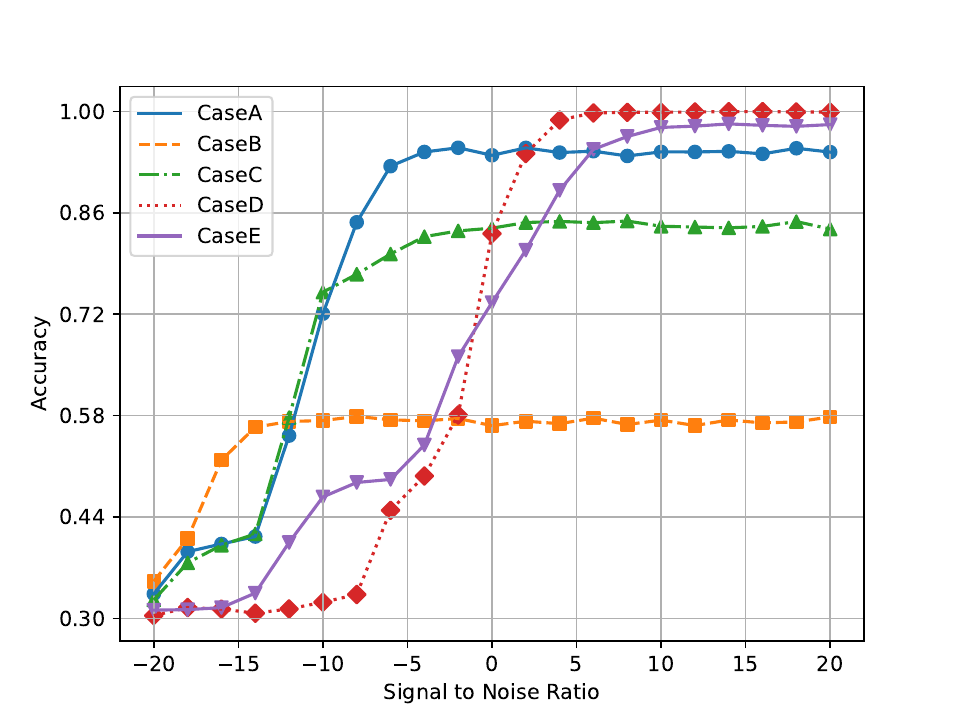}
  \vspace{-10pt}
  \caption{Classification accuracy on different SNR on different case.}
  \vspace{-9pt}
\end{figure}
\begin{figure}[h]
  \centering
  \includegraphics[width=0.5\textwidth]{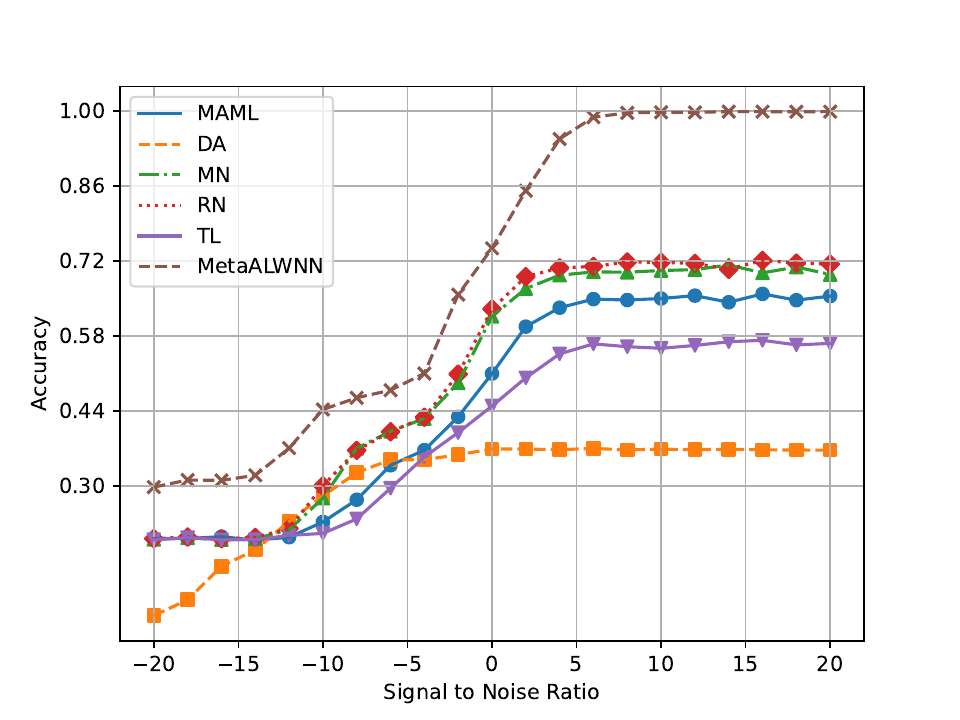}
  \vspace{-10pt}
  \caption{Classification accuracy on different SNR of MALWNN and other models on CaseA.}
  \vspace{-9pt}
\end{figure}

\begin{table*}[h]
\caption{INFERENCE TIME PER SAMPLE ON EDGE DEVICE}
\centering
\begin{tabular}{lcccc}
\toprule
Methods & \multicolumn{4}{c}{Inference time per sample on edge device (unit: s)} \\
\cmidrule(lr){2-5}
Batchsize
& 2 & 16 & 128 & 1024 \\
\midrule
ALWNN (proposed) & \textbf{\color{blue} $6.85\times10^{-3}$} & \textbf{\color{blue} $1.43\times10^{-3}$} & \textbf{\color{red} $9.84\times10^{-4}$} & \textbf{\color{red} $8.1\times10^{-4}$} \\
MCLDNN & $6.78\times10^{-2}$ & $1.74\times10^{-2}$ & $9.99\times10^{-3}$ & $7.97\times10^{-3}$ \\
CLDNN & $6.32\times10^{-2}$ & $1.69\times10^{-2}$ & $8.45\times10^{-3}$ & $6.88\times10^{-3}$ \\
MCNet &  \textbf{\color{red} $4.74\times10^{-3}$} & \textbf{\color{red} $1.2\times10^{-3}$} & \textbf{\color{blue} $10.25\times10^{-4}$} & \textbf{\color{blue} $8.40\times10^{-4}$} \\
ICAMC & $6.46\times10^{-2}$ & $1.55\times10^{-2}$ & $8.02\times10^{-3}$ & $6.12\times10^{-3}$ \\
AMCNet & $7.23\times10^{-2}$ & $2.15\times10^{-2}$ & $7.22\times10^{-3}$ & $6.56\times10^{-3}$ \\
CDSCNN & $9.71\times10^{-2}$ & $4.21\times10^{-2}$ & $9.24\times10^{-3}$ & $8.42\times10^{-3}$ \\
\bottomrule
\end{tabular}
\end{table*}
\vspace{-9pt}
\begin{figure}[h]
  \centering
  \includegraphics[width=0.5\textwidth]{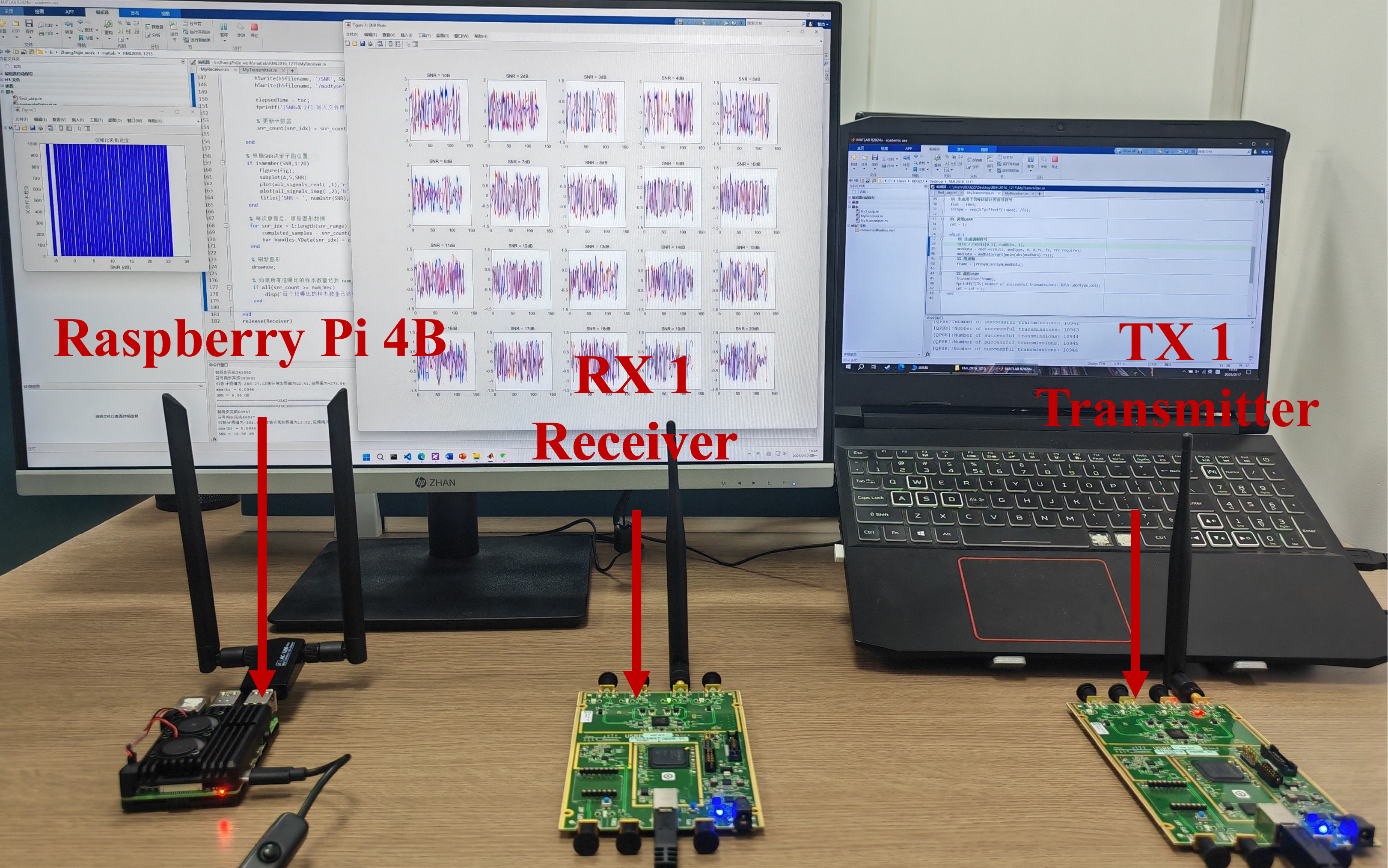}
  \vspace{-10pt}
  \caption{AMC testbed via USRP 2901s and Raspberry Pi 4B.}
  \vspace{-9pt}
\end{figure}
\vspace{5pt}
The training and test sets in CaseA were utilized to train and test the models. To ensure the reliability of the experimental results, this sample set was adopted as the data source for both the DA and TL algorithms, as well as for the algorithm proposed in this paper and the RN, MN, and MAML algorithms. When focusing on five types of modulation signals and with the sample quantity of each type precisely set at five, the variation trend of the test recognition accuracy of our algorithm and the comparison algorithms with the change of signal-to-noise ratio is vividly illustrated in Figure 8.

\section{EXPERIMENTAL RESULTS}

We conducted field experiments using USRP N2901 and Raspberry Pi 4B, evaluating the performance of our AMC method on the low-power Raspberry Pi. As shown in Figure 9, The experimental setup included two NI USRP 2901 transceivers (connected via USB 3.0 and Ethernet), a Raspberry Pi 4B, and an RTX 4090 GPU, capturing and processing real-world modulated signals such as 2ASK, FSK, AM, BPSK, QPSK, GMSK, 16QAM, and 64QAM. The test results are shown in Table V. Results showed that while ALWNN had low theoretical computational complexity, it was not the fastest model. Our proposed ULCNN exhibited limitations on the GPU but matched MCNet’s inference speed on edge devices, demonstrating better suitability for resource-constrained scenarios. Increasing batch sizes significantly reduced per-sample inference latency due to parallelized processing and optimized resource utilization. ALWNN maintained robust accuracy and practical inference speed in real-world deployments.

\section{Conclusion}
This paper proposes a lightweight AMC method ALWNN and a few-shot AMC framework MALWNN. Simulation results demonstrate that the proposed ALWNN achieves significant advantages in both accuracy and computational efficiency, while MALWNN also exhibits strong generalization capability and high precision, making them particularly suitable for edge computing scenarios with limited hardware resources and scarce training data. In future work, we will explore the implementation of AMC algorithms under multimodal data inputs.
\section*{Acknowledgement}
This work was supported by the National Key Research and Development Program of China (2020YFB2907500).

\ifCLASSOPTIONcaptionsoff
  \newpage
\fi

\bibliography{ref}
\bibliographystyle{IEEEtran}

\end{document}